\documentclass[sn-basic]{sn-jnl}

\theoremstyle{thmstyleone}%
\theoremstyle{thmstyletwo}%

\theoremstyle{thmstylethree}%

\usepackage{subcaption}
\begin{document}

\title[Deep Learning for Opinion Mining and Topic Classification of Course Reviews]{Deep Learning for Opinion Mining and Topic Classification of Course Reviews}


\author{Anna Koufakou}
\affil{\orgdiv{Department of Computing \& Software Engineering}, \\ \orgname{Florida Gulf Coast University}, \country{USA}.  E-mail: akoufakou@fgcu.edu}


\abstract{Student opinions for a course are important to educators and administrators, regardless of the type of the course or the institution. Reading and manually analyzing open-ended feedback becomes infeasible for massive volumes of comments at institution level or online forums. In this paper, we collected and pre-processed a large number of course reviews publicly available online. We applied machine learning techniques with the goal to gain insight into student sentiments and topics. Specifically, we utilized current Natural Language Processing (NLP) techniques, such as word embeddings and deep neural networks, and state-of-the-art BERT (Bidirectional  Encoder  Representations  from  Transformers), RoBERTa (Robustly optimized BERT approach) and XLNet (Generalized Auto-regression Pre-training). We performed extensive experimentation to compare these techniques versus traditional approaches. This comparative study demonstrates how to apply modern machine learning approaches for sentiment polarity extraction and topic-based classification utilizing course feedback. For sentiment polarity, the top model was RoBERTa with 95.5\% accuracy and 84.7\% F1-macro, while for topic classification, an SVM (Support Vector Machine) was the top classifier with 79.8\% accuracy and 80.6\% F1-macro. We also provided an in-depth exploration of the effect of certain hyperparameters on the model performance and discussed our observations. These findings can be used by institutions and course providers as a guide for analyzing their own course feedback using NLP models towards self-evaluation and improvement.}


\maketitle

\section{Introduction}
The course feedback from students has long been utilized from individual instructors to groups and institutions for a variety of purposes. Instructors can utilize the course feedback in order to find out what is important to students, the effectiveness of teaching material and methods, etc., in order to improve the course for a future offering. Institutions can use student surveys to gauge student perceptions and opinions, and even towards evaluations of instructors. Quantitative data from surveys, such as student ratings of their course instructors, have been used for a long time in an effort to measure teaching effectiveness: ``student ratings are the single most valid source of data on teaching effectiveness - in fact there is little support for the validity of any other source of data'' \citep{spencer2002student}.

Besides the quantitative data, there are also qualitative data in the form of textual responses, which are not as easy to explore, summarize, or visualize. Many issues exist with student text comments, such as misspellings, abbreviations, short or irrelevant statements, rambling, etc. However, because of the open-ended nature of the feedback, students are allowed to describe what is on their mind and what they feel is important without necessarily thinking of ratings or matching quantitative concepts such as a Likert scale. At the same time, the issues associated with understanding human language at scale have kept many from fully utilizing this resource. 

In the past decade or so, there has been an increasing amount of research using text-based student course feedback for a variety of tasks and purposes. With the ubiquitous web-based platforms and social media, there is also an abundance of data to collect and analyze, for example, from twitter \citep{chen2014mining} or ratemyprofessor website \citep{onan2020mining}. Many efforts are focusing on \textit{sentiment analysis}, which is the field of study that analyzes people’s opinions, sentiments, attitudes, and emotions in text.  
There has been a lot of research using sentiment analysis on educational data, for example see the surveys in \citep{dolianiti2018sentiment,zhou-2020}. Just like sentiment analysis has been used by businesses to help improve marketing and customer relationships, in the educational field, sentiment analysis may be used to improve the ``attractiveness of higher educational institutions'' \citep{santos2018improving} or decrease drop-out rates in Massive Open Online Courses (MOOCs) \citep{kastrati2021sentiment}. 

Besides sentiment analysis, there are other tasks that have been explored in related research, for example topic modeling or topic-based classification: the goal here is not to extract the sentiment (positive or negative), but to extract or predict the topics for which the comments were written or focused on, for example \citep{van2018uit,srinivas2019topic}. There is also work in aspect-based sentiment mining, which targets the sentiment for each specific entities (aspects) in the comments, for example \citep{sindhu2019aspect,ren2022automatic}.

Much of the earlier work related to course feedback analysis has concentrated on traditional Machine Learning (ML) techniques, for example \citep{altrabsheh2014sentiment,koufakou2016using}. In the last few years, researchers have taken advantage of the advances in ML and Natural Language Processing (NLP) and used Deep Learning (DL) models, for example utilizing word embeddings, and convolutional or other deep neural networks; as examples, see \citep{dessi2019evaluating,onan2020mining, estrada2020opinion}. There are also recent educational data mining surveys \citep{dutt2017systematic,kastrati2021sentiment}.

In this paper, we first describe the process of collecting and annotating a corpus of more than ten thousand online reviews for a variety of courses with topics ranging from Web Development to Data Science to Marketing. We applied sentiment polarity extraction on our corpus, looking at reviews as positive or negative. We also explored topic-based classification, categorizing each review in one of four topics. We performed extensive comparative analysis of several DL techniques (such as CNNs and LSTMs, using word embeddings, but  also state-of-the-art models, BERT, RoBERTa, and XLNet) and compared their efficacy with traditional classifiers (k-Nearest Neighbor, Na\"ive Bayes, and Support Vector Machines SVMs). Besides our extensive experimentation with very different classifiers, we also explored further possible improvements on the accuracy and the effects on runtime efficiency of the DL models. The main contributions of our work are: 
\begin{enumerate}
\item We utilized a brand new corpus we collected from over ten thousand course reviews online. Using the new corpus, we presented a two-fold analysis (opinion-based and topic-based) as opposed to many previous works who focused only on one task. This way, our work can demonstrate how to employ the data for different tasks and highlight similarities and differences between the two tasks. For example, in our experiments for topic-based classification, we found that an SVM model performed better than the DL models, which was not the case for the opinion mining experiments.
\item We utilized and experimented with, not only traditionally-used Deep Learning models, such as CNNs and RNNs, but also state-of-the-art NLP transformer-based models, namely BERT, RoBERTa, and XLNet. BERT (and as a consequence any similar models) has quickly become the de facto baseline in NLP experiments \citep{rogers-etal-2020-primer}. Our literature review in the area of course feedback analysis found only a handful of papers that used BERT and none that used RoBERTa or XLNet (see Section \ref{sec:lit-review}).
\item We reported the performance results and observations from extensive experiments with diverse models (traditional, deep learning, and transformer-based) we employed for classification. Our experimentation is rigorous and built on a solid framework, for example we used cross-validation, we reported several metrics, we compared confusion matrices, among other things.  Additionally, we explored how to improve the accuracy of our DL models, while looking at the relation of the improved accuracy on runtime. We have not seen any similar exploration in related work for course feedback analysis. 
\end{enumerate}

The organization of this paper is as follows. In Section \ref{sec:lit-review}, we review previous work related to ours. In Section \ref{sec:corpus}, we describe the corpus we developed and used in this study. In Sections \ref{sec:methods} and \ref{sec:results}, we provide a detailed view of our models followed by our experiments and results. Finally, in Section \ref{sec:conclusions}, we summarize
our work and provide concluding remarks.

\section{Related Work}
\label{sec:lit-review}
There has been an ever increasing amount of research in using text mining and NLP for educational purposes due to the increased processing power, abundance of datas and the recent advances in ML and NLP models. In the following, we review traditional techniques applied to the analysis of student course reviews, followed by current work in this area, namely using DL. 

We also provide a summary of representative related work in Table \ref{tab:table-ref-summary}: the Table lists the ML techniques and what types of data were used in each reference, listed by the first author and year of publication for space.

\subsection{Traditional techniques}

In the previous decade, there are several articles in the literature that have used traditional techniques and tools to mine student feedback. For a thorough review of earlier work, see surveys such as \citep{pena2014educational}. In the following, we give an overview of representative works in this area based on traditional techniques.

\cite{sliusarenko2013text} used key-phrase extraction and factor analysis to identify which factors were important in student comments; they also employed regression analysis to find which factors have the most impact on student ratings. \cite{altrabsheh2014sentiment} 
applied several pre-processing techniques to data they collected and then machine learning algorithms for sentiment analysis, finding that the
best method was the Linear SVM using unigrams.
\cite{ortigosa2014sentiment} performed sentiment analysis by combining lexicon-based techniques with ML models such as Decision Tree, Na\"ive Bayes and SVM. \cite{tian2014recognizing} recognized emotions (such as anger, anxiety, or joy) in Chinese texts from e-learners and proposed a framework for regulating the e-learner’s emotion based on active listening strategies.
\cite{koufakou2016using} explored sentiment analysis using traditional techniques based on bag-of-words, Na\"ive Bayes and k-Nearest Neighbor, as well as  Frequent Itemset Mining to identify key frequent terms in survey comments. As more recent examples, \cite{lalata2019sentiment} employed an  ensemble  of traditional ML algorithms, specifically, Na\"ive  Bayes,  Logistic  Regression,  SVMs,  Decision  Tree and  Random  Forest. 
\cite{hujala2020improving} used an LDA (Latent Dirichlet Allocation) method, and then applied qualitative and quantitative evaluation methods to validate the outcomes by connecting them to theoretical frameworks and quantitative data. 
 
As to the type of data that has been used, researchers have collected data e.g. from social media, or used data from their own courses. As examples, researchers have collected real-time student feedback from lectures as well as end-of-semester surveys \citep{altrabsheh2014sentiment}; survey data from courses at their department \citep{koufakou2016using}; students’ conversations on twitter to understand students’ opinions about the learning process \citep{chen2014mining};  facebook data, for example messages on the user wall and basic user data such as gender and birthday \citep{ortigosa2014sentiment}. More recent examples: \cite{hujala2020improving} used over six thousand student survey results carried out at a Finnish University; \cite{abbas2022students} used student evaluations of more than five thousand teachers from a University in Mexico. 
 
 Researchers have also shown how to use the extracted sentiment to predict student performance. For example, \cite{sorour2015correlation} applied probabilistic latent semantic analysis (PLSA) on student comments collected after specific lessons in introductory programming courses. Then, they predicted student final grades using SVM and artificial neural networks (ANN), where the SVM had the highest accuracy. 
 
\begin{table}[]
\caption{Summary of Representative Related Works. References listed as \textit{first author, year},  in chronological order. \\\footnotesize{Traditional: NB=Na\"ive Bayes, DT=Decision Trees, RF=Random Forest, ME=Maximum Entropy, kNN=k-Nearest Neighbor, LR=Logistic Regression, MLP=Multi-Layer Perceptron, LDA=Latent Dirichlet Allocation. DL: CNN=Convolutional NN, GRU=Gated Recurrent Units, LSTM=Long-Short Term Memory, RNN: Recurrent NN, BERT=Bidirectional  Encoder  Representations  from  Transformers.}
}
\label{tab:table-ref-summary}
\centering
\begin{tabular}{ll}
\hline
\textbf{Reference} & \textbf{Models; Data (language if known)}\\
\hline
Altrabsheh, 2014 
& NBs,   ME, SVM, with various pre-processing; surveys from\\
&  various universities in UK (English) \\
\hline
Chen, 2014 
& NB multi-label   classification; twitter related to engineering\\
&  students (English) \\
\hline
Ortigosa, 2014 
& Lexicon-based techniques combined  with DT, NB, SVM; \\
& various data extracted from facebook \\
\hline
Koufakou, 2016 
& kNN,   NB, frequent itemset mining; surveys from  their \\
& University (English) \\
\hline

Nguyen, 2018 
& NB,   SVM, variants of LSTMs; Vietnamese   Student Feedback \\
& Corpus UIT-VSFC (Vietnamese)\\
\hline
Tseng, 2018  
& NB,   NN, RNN, LSTM, attention;
surveys   from their \\
& University (Chinese) \\
\hline
Yu, 2018 
& SVM,   CNN with text and quantitative data; surveys   from\\
&  their University (Chinese) \\

\hline
Dessi, 2019 
& SVM,   RF, MLP with context-trained    embeddings; data \\ 
&    extracted earlier from Udemy (English) \\
\hline
Lalata, 2019 
& Ensemble with  NB, LR, SVM, DT, RF; surveys   from their \\
& University (English, Filipino) \\

\hline
Srinivas, 2019 
& LDA for topic modeling then VADER for sentiment \\
& analysis; niche.com \\
\hline

Estrada, 2020 
& NB,   SVM, kNN, DT, RF, LSTM, CNN, BERT, Evolutionaly\\
&  model; various websites such as twitter and youtube \\
& and feedback from programming related courses in their \\
& University (Spanish) \\
\hline

Onan, 2020 
&  Various embeddings (e.g. word2vec, GloVe) with NB, SVM, \\ 
& AdaBoost, CNN, LSTM, RNN, GRU; ratemyprofessor.com\\
\hline
Rybinski, 2020 
& Various   BERT models; various   websites such as \\
& ratemyprofessor, and their own University feedback (English)\\
\hline
Truong, 2020 
& PhoBERT   pre-trained BERT model for Vietnamese;     UIT-VSFC \\
& (Vietnamese)\\
\hline
Ren, 2022 & Aspect-based SA: topic/sentiment dictionaries with bi-LSTM\\
& attention; open-ended questions to junior school (Chinese) \\
\hline
\end{tabular}
\end{table}

Besides applying ML techniques to student feedback, there has also been work on developing tools and frameworks for analysis of student feedback. For example, a conceptual framework for student feedback analysis by \cite{gottipati2017conceptual} included a sentiment extraction stage and logistic regression.
\cite{gronberg2021palaute} proposed an open-source online text mining tool for analyzing and visualizing student feedback entered in course surveys at a university. \cite{estrada2020opinion} proposed an emotion recognition and opinion capture as part of an integrated learning environment for Java. \cite{srinivas2019topic} used LDA for topic modeling and the Vader tool (proposed by \cite{hutto2014vader}) for sentiment classification, as part of a larger systematic view on strengths, weaknesses, etc. to be used by universities to analyze online feedback.

\subsection{Deep Learning (DL)}
 
DL is a more recent advancement in the larger field of machine learning and it has already been used for educational data mining successfully \citep{doleck2020predictive}. DL models are usually Artificial Neural Networks (ANNs) with more layers than traditional ANNs. They include networks such as Convolutional Neural Networks (CNN),  Long Short-Term Memory (LSTM), etc. Among the more recent neural architectures, BERT \citep{devlin-etal-2019-bert} is considered state-of-the-art in several NLP tasks. 
We review the basics of the models we use in our work in Section \ref{sec:methods}.

For a thorough view on the recent work related to the analysis of student feedback, the reader is referred to recent related surveys, such as \citep{dutt2017systematic, dolianiti2018sentiment, kastrati2021sentiment}. The most recent survey we found \citep{kastrati2021sentiment} shows that there are only seven  papers related to this topic that utilized DL methods, even though we found additional works dated after the survey, using DL  models. In the following paragraphs, we review representative work that is related to our work.

\cite{yu2018improving} used sentiment information extracted from student self-evaluations to improve the accuracy of early prediction of which students are likely to fail in a Chinese course. They used a Chinese affective lexicon, and structured data such as attendance, in conjunction with unstructured text comments. They found that CNNs using both structured and unstructured data had the best performance overall. Another study by \cite{tseng2018text} also focused on course surveys with the task to use student comments for evaluating and hiring teaching faculty. They compared deep networks such as Recurrent Neural Networks (RNNs) using a Chinese text sentiment analysis kit, named SnowNLP, and they found that the best accuracy was achieved by an attention LSTM classifier. 

There is a branch of related work based on Vietnamese data. \cite{van2018uit} developed a Vietnamese Students’ Feedback Corpus named UIT-VSFC, human-annotated for classification based on sentiment and on topics. \cite{nguyen2018variants} explored variants of LSTMs for sentiment analysis on that corpus. 
\cite{truong2020sentiment} utilized PhoBERT, a pre-trained BERT model for Vietnamese, and fine-tuned it to achieve state-of-the-art results on UIT-VSFC.

\cite{dessi2019evaluating} experimented with several word embedding representations and DL as well as traditional models, for regression based on a sentiment score rating. They found that the best performance is achieved by their Bidirectional LSTM with an attention layer, based on word2vec. They also explored training word embeddings on a relevant corpus (not available as far as we see).

\cite{estrada2020opinion} presented sentiment analysis and emotion detection of online data from various sources, such as youtube or twitter, as well as data collected from their own courses. They utilized different models such as CNN and LSTM, as well as BERT and an evolutionary model; the latter performed the best in their experiments. Their DL models used as input one-hot encodings of the student comment text, and not word embeddings. \cite{onan2020mining} also focused on sentiment analysis and experimented with various embeddings (word2vec, GloVe, fastText, LDA2Vec) and models such as CNN, RNN, LSTM, as well as ensemble techniques. Their experimentation is very thorough and they used more than a 150 thousand reviews collected from ratemyprofessor website. 

There is also work that has focused on \textit{aspect-based} sentiment analysis, which targets sentiment related with a specific aspect, for example, instructor or course. \cite{sindhu2019aspect} applied a two-layer LSTM with the goal of aspect-based sentiment analysis on their own University data as well as SemEval-2014 data. The first layer predicted aspects from the feedback, while the second predicted the sentiment polarity. 
\cite{kastrati2020weakly} used more than a 100 thousand reviews from coursera as well as classroom feedback. They applied LSTMs and CNNs  using various word embeddings. \cite{ren2022automatic} used Chinese open-ended comments written by junior school students. They constructed dictionaries for topics and sentiments which were used for their deep learning model to predict sentiments.

From our review, the research in analysis of student feedback has not fully embraced the state-of-the-art models, namely BERT and its extensions such as RoBERTa or XLNet. To start with, we already cited \citep{estrada2020opinion, truong2020sentiment} earlier in this section. \cite{rybinski2020will} compared BERT models on 1.6 million student evaluations from the US and the UK, extracted from different sources. \cite{wang2020makes} used subtitles (captions) in videos of more than thousand courses to predict the performance of the instructor in online education, using a hierarchical BERT model based on teacher's verbal cues and on course structure.

In summary, even though there has been a considerable amount of research work in student feedback analysis, there is still a gap of utilizing the recent state-of-the-art models such as BERT, which our paper aims to fill. In our review, we noted that most of the work focuses  on sentiment analysis, while our work also examines topic classification. We also noted that several works did not report metrics other than accuracy (for example, \citep{estrada2020opinion}) or they did not describe their DL models or (hyper)parameters (for example, \citep{tseng2018text}). We presented extensive experimentation for two different classification tasks, with various DL models, exploring the effect of hyperparameters, and discussed runtime efficiency as well as classification accuracy. 

\section{Dataset Description}
\label{sec:corpus}
For this work, we collected publicly available course reviews posted online for bootcamp-type courses at website https://www.coursereport.com.
The reviewed courses were for various topics, from assembly language, to web development, to marketing, and the courses were offered online or in different cities globally. The vast majority of the reviews were in English. The data contained a course title, a course review (text comments), a review rating (1 through 5), and other fields we did not use, for example, username or instructor rating or helpfulness of the review. We used a web crawler we developed from scratch to collect data for the reviews. We pre-processed the text reviews to clean up invalid text: removed remaining HTML tags and any reviews that are shorter than 2 words. The resulting dataset had 10,610 reviews (text comments). The review text length ranges from a minimum of 2 words to a maximum of 4,219 words, with an average of 245 words and a standard deviation of 251.

\begin{figure}[t] 	
\centering 	
\includegraphics[scale=0.5]{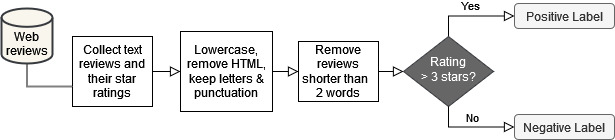}
\caption{The overall pre-processing and labeling tasks for the sentiment analysis task} 	
\label{fig:fig-preprocess} 
\end{figure}

\begin{table}[b]
\centering
\caption{Example comments from course reviews in our data and their label}
\begin{tabular}{ll}
\hline
\textit{\textbf{Label}} & \textit{\textbf{Sample from a course review} }\\
\hline
Positive & ``I would not hesitate to recommend $<$program$>$'' \\
& ``Great curriculum and instructors ... they make sure you are 
prepared'' \\
& ``$<$name$>$ is an awesome teacher and all of the TAs were
very helpful'' \\
& ``The $<$course$>$ was worth the money and gave me a good  insight into\\
& ~~~ what the $<$program$>$ would be like'' \\
\hline
Negative & ``This is a terrible program. It was very badly organized.''  \\
& ``This is one of the worst, most expensive programs in [...]'' \\
& ``None of my submissions so far have been reviewed or have any feedback \\ 
& ~~~ [...] with 2 weeks left in the program'' \\
& ``I feel $<$program$>$ at $<$school$>$ was a waste of 
time and money''\\ 
\hline
\end{tabular}
\label{tab:table-example-reviews}
\end{table}
First, we organized the data for the sentiment polarity extraction task, as follows. As mentioned above, each review had a star rating, ranging from 1 to 5. The dataset was divided into positive reviews and negative reviews as follows: the positive comments were considered the ones which had rating score 4 or 5, while the negative comments had score ratings 1-3. The entire dataset ended up very imbalanced: it contained 91.5\% positive and 8.5\% negative reviews. Fig. \ref{fig:fig-preprocess} shows the pre-processing and labeling steps. Table \ref{tab:table-example-reviews} shows examples of comments taken from positive and negative reviews. 

\begin{table}[b]
\centering
\caption{Top Ranking Features (trigrams) for positive and for negative course reviews} 	
\begin{tabular}{ll}
\hline
\textit{\textbf{Label}}    &\textit{\textbf{ Top Ranking Features}}    \\
\hline
Positive & web development course,  short period time,  \\   
&         did great job,  life changing experience, \\
         & highly recommend course, make career change\\
         \hline
Negative ~~~& waste time money, web development course,  \\
 &        don't waste time, free online resources,  \\
         & don't waste money, job placement assistance   \\
         \hline
\end{tabular}
\label{tab:table-trigram} 
\end{table}

\begin{table}[b]
\centering
\caption{Topics, Example Courses in each Topic, Distribution (percentage) of each topic, and total number of comments in the Topics Dataset}
\begin{tabular}{cllr}
\hline
 \textit{\textbf{Label}} & 
 \textit{\textbf{Topic / Category}}       & 
 \textit{\textbf{Example Courses in the Topic }} & \multicolumn{1}{c}{\textit{\textbf{\%}}} \\ \hline
1 & Programming            & .NET, iOS, Java                  & 33.20  \\
2 & Web Development        & Web Development, JavaScript      & 50.74  \\
3 & Non-Programming        & UX Design, Marketing             & 9.61  \\ 
4 & Data Science/Analytics & Data Science, Bus. Analytics & 6.45   \\ 
\hline
\end{tabular}
\label{tab:table-topics-data}
\end{table}

Finally, we examined the top ranking trigrams for the positive versus the negative reviews, shown in Table~\ref{tab:table-trigram} (we also looked at unigrams and bigrams but they were not as descriptive of the polarity between the reviews as the trigrams). In Table~\ref{tab:table-trigram}, one can see that ``web development course'' is a top ranking term for both types of reviews. We observed that this course is the most frequent course topic overall, therefore it appears very frequently in both negative and positive reviews. Other terms such as ``waste time money", ``dont waste time", ``free online resources" rank high in negative reviews, while positive reviews have terms such as ``highly recommend course" and ``life changing experience". 

For the topic-based classification, first we looked at the course titles and their reviews, and we conducted different visualizations, for example, word clouds. Additionally, we utilized  Latent Dirichlet Allocation (LDA)\citep {blei2003latent} and we identified the major course topics were, by far, Web development, Programming, and Data Science. We also manually filtered reviews based on similar course titles and then grouped together these courses in one topic or category. Finally, we dropped the rest of the reviews that did not have any course name or identifiable topic, and ended up with 7,503 reviews. The topics and their distribution in the resulting data is shown in Table~\ref{tab:table-topics-data}. As shown in Table~\ref{tab:table-topics-data}, the large majority of the courses are related to programming or web development.

\begin{figure}[b] 	
\centering 	
\includegraphics[scale=0.5]{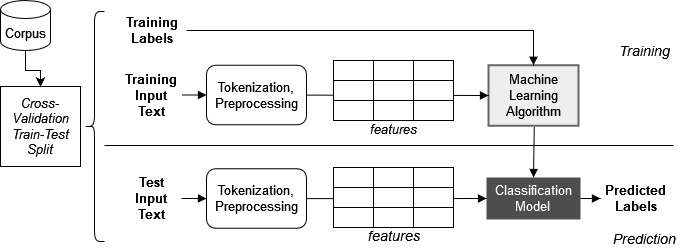}
\caption{The training and prediction process for classification using a machine learning algorithm} 	
\label{fig:fig-model} 
\end{figure}

\section{Methodology}
\label{sec:methods}
The general process for classification using an ML algorithm in our work is shown in Fig. \ref{fig:fig-model}. The data is split into training and test set, each including the text and the labels (for example, positive or negative). The training data is used to extract the \textit{vocabulary}: the set of unique tokens or words found in the training data. Based on the vocabulary, we then extract the features that train the model (more details on the features will be given in the following sections). In the prediction phase, the model generated from training is then used to predict the labels for the test inputs. For the split of the dataset into training and test sets, we used cross-validation (see Section \ref{sec:exp-setup}). In \textit{k}-fold cross-validation, the data is split into a total of \textit{k} subsets, then the experiment is executed \textit{k} times, ensuring the test set is varied in each execution.  

In the rest of this section, we first present an overview of a traditional approach for classification of text, based on Bag-of-Words (BoW). We also briefly present the BoW classifiers we use in our experiments. Then, we give an overview of DL models for text classification in general, as well as the specific models we use in our work.

\subsection{Traditional Bag-Of-Words (BoW) Approach}

First, we extracted the text from our collection of course reviews (corpus) and then tokenized the text into words (see Section \ref{sec:corpus} for more details on pre-processing). The resulting dataset was represented as a BoW matrix. In these methods, the features extracted from the data in Fig. \ref{fig:fig-model} is the BoW matrix. 

BoW methods do not preserve the order of the words in the text or the context of a word in a phrase, nor do they preserve or extract grammar-related or other relations: they only store frequency information for each unique word in the corpus. The dimensionality of the resulting BoW matrix is the number of documents (or unique course reviews) $\times$ the unique words or tokens in our corpus (the vocabulary). A small detail is that the vocabulary is extracted from the training set only. 

For this part of our work, we used TF-IDF (Term Frequency–Inverse Document Frequency) values. TF-IDF is a statistical measure used to evaluate the importance of a word in a document in a corpus.  
Using  TF-IDF, the importance increases proportionally to the frequency of the word in the document, but it is offset by the frequency of the word in the corpus. We used the TF-IDF features as input to three classification models with the goal either to predict if a course review comment is negative
or positive (sentiment analysis) or to detect one of the four topics (topic-based classification): Naïve Bayes, k-Nearest Neighbor (k-NN), and Support Vector Machines (SVM). These algorithms are briefly described below - for more details see any related text, such as \citep{tan2005introduction}. 
 
Naïve Bayes offers a probabilistic framework for solving classification problems. Naïve Bayes first uses the training data (corpus) to find the probability of each unique word as it occurs in the corpus for each class. For a test document, Naïve Bayes multiplies the pre-calculated probabilities of every word in the document and then chooses the class with the highest probability to classify the test record.

In the $k$-Nearest Neighbor algorithm, given a course review $x$ and a user parameter $k$, the algorithm finds the $k$ reviews that are the most similar to $x$. These are called its $k$-nearest neighbors. Then, based on the majority of the labels of $x$’s $k$-nearest neighbors, the algorithm predicts the label of $x$. 

SVMs have been very successfully applied to a number of applications since their inception, including analysis of course feedback (see Section \ref{sec:lit-review}). The SVM algorithm finds a hyperplane (decision boundary) that separates the classes by their features over a space \citep{cortes1995support}. The goal is to maximize  the margin, or create  the  largest  possible distance  between  the  separating  hyperplanes,  in order to  reduce  the  upper  bound  on  the expected  generalization  error. For non-linearly separable data, the solution is to map the inputs into a high-dimensional feature space. 

\subsection{Deep Learning (DL) Approach}

\subsubsection{Word Embeddings}
As already mentioned in the previous section, traditional methods for representing words in matrix form, such as BoW and TF-IDF, do not take into account the position or the context of the word in the document. Recent approaches proposed word embeddings that represent the semantic meanings of words \citep{mikolov2013distributed}. Words that are similar in meaning or in context are closer to each other in the vector space, while words that are different are farther apart. In this work, we experimented with Word2Vec \citep{mikolov2013distributed}, which uses a feed-forward neural network to predict the neighboring words for a given word in order to create the word embeddings. The embedding for each word is essentially a one-dimensional vector of $d$ values, where $d$ is a user-entered parameter. 

\subsubsection{Deep Neural Networks}
In contrast to traditional techniques in Section \ref{sec:methods}.1, more recent approaches use DL, such as Convolutional Neural Networks (CNN) and Recurrent Neural Networks (RNN), to learn text representations. In the following, we give a brief background review on the models we used in our work; the reader is referred to  \citep{minaee2021deep} for a comprehensive review on DL-based Text Classification. We also list the models we used for this work, and provide a detailed step-by-step example for a convolutional model we employed for our work.

\begin{figure}[t] 	
\centering 	
\includegraphics[scale=0.49]{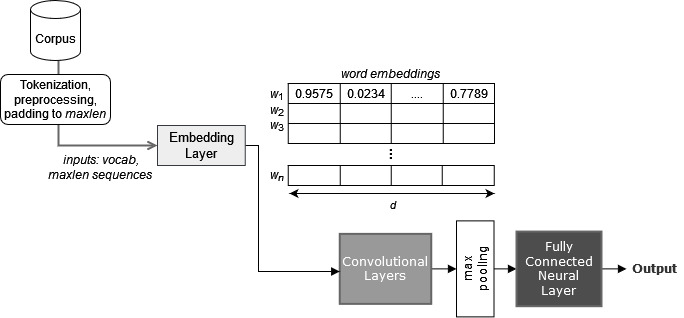}
\caption{A depiction of a CNN-based model we used for classification} 	
\label{fig:fig-deep-models} 
\end{figure}

Originally invented for computer vision, CNN models have subsequently been shown to be effective for many NLP tasks \citep{kim2014convolutional}. CNNs utilize layers with convolving filters. In text related tasks, the filters are trained to identify word combinations that are most pertinent to the classification task at hand. In most of the recent literature, the word embeddings from the document are fed into the NN as features. In addition, as character-based CNNs have been shown to work for text related tasks \citep{zhang2015character}, we briefly experimented with a character-based CNN.

RNNs were also shown to be effective in NLP tasks due to their architecture specifically designed to address time-series data. In NLP tasks, RNNs are aiming to learn linguistic patterns based on different sequences of words. Basic RNNs are unable to retain information and find relationships over a large sequence of words so we used LSTMs: Long Short-Term Memory units (LSTM) \citep{hochreiter1997long} use gating functions to selectively store or “forget” input information according to how relevant it is to the classification task. Finally, we also experimented with a Bidirectional LSTM model, which ensures that the network can account for the \textit{preceding} as well as the \textit{following} context when processing the sequence of words.

An example of a convolutional model we employed based on word embeddings is shown in Fig. \ref{fig:fig-deep-models}. As before, the collection of documents or course reviews was tokenized into words, but now we padded or shortened each resulting review to a set length (or number of words), based on a user-entered parameter called \textit{maxlen}. We extracted the vocabulary from the resulting text sentences, and then created word embeddings of length $d$. The  result from the embedding layer was a three-dimensional matrix, of dimensionality $ n \times \textit{maxlen} \times d$, where $n$ is the vocabulary size and $d$ is the embedding dimension. 

The embeddings were fed into the convolutional layer. The output of the convolutional layer was fed into a dropout and a max-pooling layer. There might be more than one convolutional layer employed in this model; if so, the outputs were concatenated before the next layer. Finally, we used a fully-connected dense layer to output the prediction of the model (the predicted label). This layer used a \textit{sigmoid} or a \textit{softmax} activation, depending on the label being binary (sentiment) or categorical (topic), respectively. Our LSTM or bi-LSTM models follow a similar general idea. 

\subsubsection{Transformer-based models} 
While word embeddings take into consideration the semantic similarities of words in a corpus, they do not explore different meanings of words based on context. Therefore, in word embeddings such as word2vec \citep{mikolov2013distributed}, each word in the vocabulary will have one single embedding. More recent techniques introduced \textit{contextualized} embeddings: they encode a word and its context from the words before it, and after it, so it ``will generate a different embedding vector for the word `bank' in `bank account' to that for `river bank' '' \citep{rybinski2020will}. 

BERT (\textit{Bidirectional  Encoder  Representations  from  Transformers}) \citep{devlin-etal-2019-bert} is considered state-of-the-art in several NLP tasks. For example, in a recent SemEval Task for detecting offensive language \citep{
zampieri2020semeval}, the vast majority of the top entries in the task used BERT-like systems. A recent survey found that ``in a little over a year, BERT has become a ubiquitous baseline in NLP experiments'' \citep{rogers-etal-2020-primer}. A transformer combines a multi-head self-attention mechanism with an encoder-decoder. BERT utilized Masked Language Modeling (MLM) and Next Sentence Prediction (NSP). In MLM, the model masks some of the words, and uses the rest of the words to predict the masked words. In NSP, given two sentences, BERT was trained to predict if the second sentence is likely to follow the first sentence. For more information on the \textit{internal} architecture of BERT, the reader is referred to \citep{devlin-etal-2019-bert}.

Utilizing BERT is somewhat similar to the models such as CNN we discussed in the previous section, with some significant differences. One of these differences is the BERT \textit{tokenizer}. The BERT model needs inputs in the form of: token identifiers, masks, and segments. BERT marks the end of each sentence with a special [SEP] token. Also, BERT inserts a [CLS] token (which stands for “classification”) to the start of each sentence. 

Besides these, the BERT-based model we employed is overall similar to the previous CNN model in Fig. \ref{fig:fig-deep-models}. The BERT-based model also uses a \textit{maxlen} for the input comments (see the definition of \textit{maxlen} in the previous section for the CNN model, and Fig. \ref{fig:fig-deep-models}). The results from the BERT tokenizer were passed onto the BERT layer, whose [CLS] output was fed into a dense layer that outputs the prediction of the model. Just as in the DL models from the previous section, this layer used a \textit{sigmoid} or a \textit{softmax} activation, for binary (sentiment) or categorical (topic) classification, respectively. 

An important benefit of using a BERT-based model, over a CNN-based model such as the one in Fig. \ref{fig:fig-deep-models} is the BERT model has already been pre-trained on big data: in fact, many pre-trained models exist available for direct use or that can be fine-tuned for a specific classification task. Fine-tuning means to further train the pre-trained BERT model using our data. Section \ref{sec:exp-setup} includes the details of our BERT model and its hyperparameters. 

There have been numerous models extending BERT. In our experiments, we used RoBERTa and XLNet. As main differences from BERT, RoBERTa (Robustly optimized BERT approach) \citep{liu2019roberta} removed the NSP and replaced the static masking (in MLM) of BERT, with dynamic masking. In summary, RoBERTa has been shown to be more robust than BERT, it modified some of BERT, and it was trained using more data. XLNet \citep{yang2019xlnet} was based on an auto-regressive model,  which predicts future behavior based on past, and used a Transformer-XL. XLNet also introduced permutation language modeling, where \textit{all} tokens (not \textit{only} masked tokens) are predicted in random order, rather than sequential.

\section{Experiments and Results}
\label{sec:results}

\subsection{Experimental Setup}
\label{sec:exp-setup}
As discussed
in Section \ref{sec:corpus}, the dataset we collected for the sentiment analysis is very imbalanced: it contains 91.5\%
positive and 8.5\% negative reviews. Therefore, for our sentiment-based classification, we used stratified 10-fold cross validation (CV).  
For the topic-based classification, we used stratified 5-fold CV to better suit the four topics and their distribution (see Table \ref{tab:table-topics-data}).

In order to implement the diverse suite of classification models we utilized, we wrote our code using different tools and platforms, which resulted in quite different implementations\footnote{We plan to share our implementations on request when this paper is published}. We conducted all experiments using google colaboratory\footnote{\url{https://colab.research.google.com}}. We used scikit-learn\footnote{\url{https://scikit-learn.org}} for all our BOW experiments and
keras\footnote{\url{https://keras.io}} for our implementations of the DL models. All approaches based on TF-IDF were run with the scikit-learn
defaults and unigrams. For the NN experiments we used 5 epochs, 0.01
learning rate, 32 batch size, Adam optimizer, and 0.5 dropout. For the convolutional layer, we used rectified linear units (ReLU), and filter windows of 3, 4, or 5. For LSTMs, we used 64 units. For character embeddings, we used 16 as the embedding dimension.

For our experiments with word embeddings, we first used the publicly available
word2vec vectors that were trained on 100 billion words from
Google News\footnote{\url{https://code.google.com/archive/p/word2vec}}. The vectors had dimensionality of 300 and
were trained using the continuous bag-of-words (CBOW) architecture 
\citep{mikolov2013distributed}. Words not present in the set of pre-trained words were initialized randomly. In our results, the models that used these pre-trained vectors are denoted as ``Pre-trained". We also experimented with non pre-trained word vectors, i.e. vectors that were randomly initialized. 
 
Finally, for the experiments with the transformer models, we used Pytorch and the corresponding models provided by HuggingFace\footnote{\url{https://huggingface.co/transformers}}. Specifically, we used the \texttt{bert-base-uncased}, the \texttt{roberta-base} and the \texttt{xlnet-base-cased} models, all with lower case option. Based on performance in early experiments and following the recommendations by the original developers of BERT \citep{devlin-etal-2019-bert}, our transformer models used learning rate of $2e^{-5}$, 3 epochs, batch size 32, and \textit{maxlen} of 50. 

We reported our results based on the classification metrics defined below:

\begin{equation}
Precision = \frac{TP}{TP+FP}
\end{equation}
\begin{equation}
Recall = \frac{TP}{TP+FN}
\end{equation}
\begin{equation}\label{eq:acc}
Accuracy = \frac{TP+TN}{N}
\end{equation}

\begin{equation}\label{eq:f1}
F1\textnormal{-}score=\frac{2 \times Precision \times Recall}{Precision+Recall}
\end{equation}

\noindent where $TP$ is True Positives, $FP$ is False Positives, $FN$ is False Negatives, and $N$ is the total number of records. Besides Accuracy in Eq. \eqref{eq:acc}, we chose to also report the F1-macro which averages the F1-score in Eq. \eqref{eq:f1} over the classes: the macro-averaged F1 is better suited for showing algorithm effectiveness on smaller categories \citep{altrabsheh2014sentiment,kastrati2021sentiment}, which is important as we are working with imbalanced datasets.

\subsection{Results and Discussion}
\subsubsection{Sentiment Analysis}
\label{sec:results-senti}
The results for the sentiment analysis task are shown in  Table~\ref{tab:table-sentiment}. As shown in Table~\ref{tab:table-sentiment}, the transformer-based models performed the best overall: RoBERTa is the top performing model at 95.5\% accuracy and 84.7\% F1-macro, while BERT and XLNet follow with F1-macro of about 83\%. From the rest of the DL models, CNNs performed the best with the CNN using the pre-trained word embeddings performing the best at 92.1\% accuracy and 82.4\% F1-macro. From the TF-IDF models, accuracies were high but F1-macro results are low, around 50\% for all the models we utilized.

\begin{table}[t]
	\caption{Results for Sentiment Analysis}
	\centering
	    \begin{tabular}{llrr}
	\hline
	&	\textbf{\textit{Classifier}}	&\multicolumn{1}{c}{\textbf{\textit{Accuracy}}}	&	
\multicolumn{1}{c}{\textbf{\textit{F1-macro}}}	\\
	\hline

Traditional &	k-Nearest Neighbor	&	0.902 ~\small{±0.01}	&	0.554 ~\small{±0.06}	
\\
(TF-IDF) 	&	Naïve Bayes	&	0.907 ~\small{±0.00}	&	0.476 ~\small{±0.00}	
\\
Models		&	Linear SVM	&	0.908 ~\small{±0.01}	&	0.537 ~\small{±0.05}	
\\
	&	RBF SVM	&	0.907 ~\small{±0.01}	&	0.476 ~\small{±0.00} 	
\\
	&	Polynomial SVM	&	0.907 ~\small{±0.01}	&	0.476 ~\small{±0.01}		
\\
	\hline

Deep Learning	&	Character-based CNN	&	0.750 ~\small{±0.03}	&	0.591 ~\small{±0.02}	
\\
Models &	Word CNN	&	0.900 ~\small{±0.02}	&	0.708 ~\small{±0.05}	
\\
	&	Pre-trained Word CNN	&	0.921 ~\small{±0.02}		&	0.824 ~\small{±0.02}	
\\
	&	Word LSTM	&	0.788 ~\small{±0.03}	&	0.660 ~\small{±0.03}	
	\\
	&	Pre-trained Word LSTM	&	0.771 ~\small{±0.03}	&	0.637 ~\small{±0.05}	
	\\
	&	Word bi-LSTM	&	0.726 ~\small{±0.05}	&	0.597 ~\small{±0.04}	
	\\
	&	Pre-trained Word bi-LSTM	&	0.801 ~\small{±0.07}	&	0.674 ~\small{±0.05}	
	\\
 \hline
 
Transformers	&	BERT	&	0.953 ~\small{±0.01}	&	0.832 ~\small{±0.04}	
	\\
& RoBERTa & \textbf{0.955} ~\small{±0.01} & \textbf{0.847} ~\small{±0.03} 
\\
& XLNet & 0.952 ~\small{±0.02} &  0.834 ~\small{±0.06} \\

	\hline
	\end{tabular}
	\label{tab:table-sentiment}
\end{table}

We also performed a comparison based on \textit{maxlen} input values for BERT, RoBERTa, and XLNet: see Table~\ref{tab:table-sent-maxlen} for the sentiment analysis task. Higher \textit{maxlen} values mean using more words from each comment (or a larger part of the comment) as input to the model. As shown in Table ~\ref{tab:table-sent-maxlen}, all transformer models took about 3-4 minutes per epoch for \textit{maxlen} equal to 100 versus under 10 seconds for the word CNN. Among the three models, RoBERTa was slightly faster than BERT at under 3 minutes per epoch and XLNet was the slowest at almost 4 minutes per epoch. This means that each of the transformer models needed about 1.5-2 hours total runtime given 3 epochs and 10-fold stratified CV. We were not able to run transformer model experiments with \textit{maxlen} larger than 100 due to these long runtimes.  
XLNet and RoBERTa did the best in these experiments (about 97\% accuracy and 89\% F1-macro for \textit{maxlen}=100). 

Table~\ref{tab:table-sent-maxlen} shows that the CNN model also gained from an increase in \textit{maxlen} while its execution is under 10 seconds per epoch. Given this observation, we also 
ran experiments with CNN and \textit{maxlen} higher than 100. The resulting plot is shown in Fig. \ref{fig:maxlen}a (using regular word embeddings). As the figure shows, there was no gain for this model and the sentiment analysis task by increasing the \textit{maxlen}  to more than 150, and its F1-macro stayed at 80\%.

Overall, for the sentiment analysis task, we see the superiority of the transformer-based models, BERT, RoBERTa, and XLNet, especially how well these models perform with imbalanced data. We also see that using more words as input to the models increases accuracy at the expense of further increasing their execution time. 
Another avenue we leave for future research is to use a model such as DistilBERT \citep{sanh2019distilbert}, a distilled and smaller version of BERT that has been shown to be much faster. 

Finally, we examined records on which two of the top performing models, BERT and RoBERTa, disagreed on their predictions (given the same train/test split of the records). In the following, we provided a couple of reviews as examples (we edited the reviews for length and content, but made sure to preserve the spirit of each review). The following was correctly classified as \textit{positive} by BERT but \textit{negative} by RoBERTa: \textit{``I went in not knowing much more than how to write a simple program, and I got a good job [...] I had to spend a lot of time outside the classes to learn [...] curriculum is alright, a bit scattered [...] overall they're competent and do a good job [...]''}. Even though this review had a high star rating, as a whole it contained somewhat mixed opinions and negative wording. As a second example, RoBERTa labeled this review correctly as \textit{positive}, while BERT labeled it as \textit{negative}: \textit{``I'd like to respond to the one  review trashing X. It's totally wrong. X is passionate and smart, but if he sees you doing something you shouldn't, he is not afraid to call you out. [...] I can honestly affirm that $<$course$>$ was life changing [...]''.} The review had an argumentative tone trying to defend an instructor, and then it was very positive of the instructor and the course. Both examples show the complexity of capturing sentiment at the paragraph (or essay) level, as  researchers have previously noted \citep{ren2022automatic}. Future research could focus on the sentence level as a way of improving the sentiment classification of such reviews.  

\begin{table}
\caption{Comparison of Deep Learning models given different \textit{maxlen} values for Sentiment Analysis - Runtime is Seconds per Epoch}
\centering
\begin{tabular}{lcccc}
\hline
\textit{\textbf{Model}}   & \textbf{\textit{maxlen}} &\textit{ \textbf{Runtime}} & \textit{\textbf{Accuracy} }           & \textit{\textbf{F1-macro}}   \\ 
\hline
BERT    & 50     & 115     &  0.953 ~\small{±0.01}    & 0.832 ~\small{±0.04}   \\
BERT  & 100    & 188     &  0.958 ~\small{±0.01}    & 0.852 ~\small{±0.04}   \\
\hline
RoBERTa & 50     & 96     &  0.955 ~\small{±0.01}   &  0.847 ~\small{±0.03} \\
RoBERTa & 100    & 175     &  0.966 ~\small{±0.01}   & 0.889 ~\small{±0.04}   \\
\hline
XLNet & 50     &   123  &  0.952 ~\small{±0.02} &  0.834 ~\small{±0.06} \\
XLNet & 100    &   237  &   0.969 ~\small{±0.01} &  0.889 ~\small{±0.05}   \\
\hline
Word-Based CNN     & 50     & 5       & 0.900 ~\small{±0.02}   & 0.708 ~\small{±0.05}   \\
Word-Based CNN     & 100    & 8       & 0.924 ~\small{±0.04}   & 0.770 ~\small{±0.06}   \\ 
\hline
\end{tabular}
	\label{tab:table-sent-maxlen}
\end{table}

\subsubsection{Topic-Based Classification}
The results for the topic-based classification task are shown in Table \ref{tab:table-topics}. For this task, a linear SVM was the top performing model at 79.8\% accuracy and 80.6\% F1-macro. The transformer-based models, BERT, RoBERTa, and XLNet, were closely behind at low to high 70's for accuracy and F1-macro. The rest of the DL models were in the low to mid 60's for accuracy. The highest performing among them was the CNN model using the pre-trained word embeddings, with 65.9\% accuracy and 65\% F1-macro.

\begin{table}
	\caption{Results for Topic-Based Classification}
	\centering
\begin{tabular}{llrr}
	\hline
	&	\textbf{\textit{Classifier}}	&\multicolumn{1}{c}{\textbf{\textit{Accuracy}}}	&	
\multicolumn{1}{c}{\textbf{\textit{F1-macro}}}	\\
	\hline
Traditional 	&	k-Nearest Neighbor	&	0.735 \small{±0.07}	&	0.725 \small{±0.07}	\\
(TF-IDF) 	&	Naïve Bayes	&	0.562 \small{±0.04}	&	0.257 \small{±0.05}	\\
Models	&	Linear SVM	&	\textbf{0.798} \small{±0.04}	&	\textbf{0.806} \small{±0.05}	\\
	&	RBF SVM	&	0.797 \small{±0.05}	&	0.789 \small{±0.06}	\\
	&	Polynomial SVM	&	0.732 \small{±0.04}	&	0.668 \small{±0.06}	\\
	
\hline
Deep Learning & Word CNN	&	0.645 \small{±0.04}	&	0.623 \small{±0.07}	\\
 Models & Pre-trained Word CNN	& 0.659 \small{±0.06}	& 0.650 \small{±0.06}\\   
 &	Word LSTM	&	0.612 \small{±0.05}	&	0.583 \small{±0.06}	\\
& Pre-trained Word LSTM &	0.620 \small{±0.03}	&	0.581 \small{±0.05}	\\
	&	Word bi-LSTM	&	0.619 \small{±0.06}	&	0.583 \small{±0.07}	\\
	& Pre-trained Word bi-LSTM& 0.620 \small{±0.04}	&	0.588 \small{±0.04}\\
 \hline
Transformers	&	BERT	&	0.772 \small{±0.03}	&	0.774 \small{±0.05}	\\
& RoBERTa & 0.749 \small{±0.07} & 0.747 \small{±0.09} \\
& XLNet & 0.724~\small{±0.08} & 0.722~\small{±0.09} \\
	\hline
\end{tabular}
	\label{tab:table-topics}
\end{table}

\begin{figure}[t] 	
\centering 	
\begin{subfigure}{0.45\textwidth}
         \centering
\includegraphics[scale=0.44]{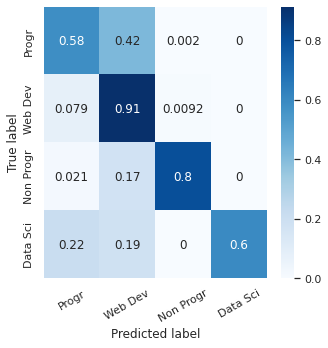}
\caption{Linear SVM Model} 	
\label{fig:fig-conf-svm} 
\end{subfigure}
\hfill
\begin{subfigure}{0.45\textwidth} 	
         \centering
\includegraphics[scale=0.44]{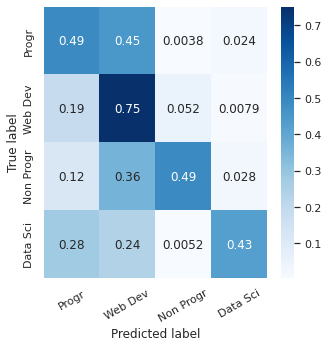}
\caption{CNN Model (regular embeddings)} 	
\label{fig:fig-conf-cnn} 
\end{subfigure}
\hfill
\begin{subfigure}{0.45\textwidth} 	
         \centering
\includegraphics[scale=0.44]{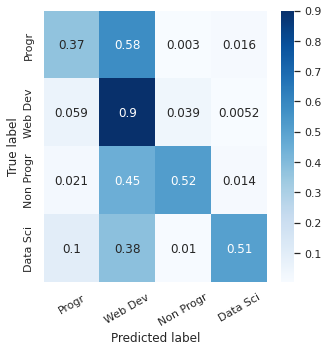}
\caption{CNN Model (pre-trained)} 	
\label{fig:fig-conf-pretrain-cnn} 
\end{subfigure}
\hfill
\begin{subfigure}{0.45\textwidth} 	
  \centering
\includegraphics[scale=0.44]{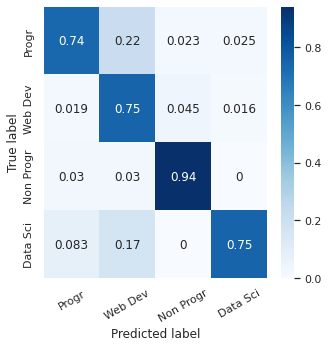}\\ 
\caption{BERT Model} 	
\label{fig:fig-conf-bert} 
\end{subfigure}
\caption{Confusion matrix for four models on  Topic-Based Classification} 	
\label{fig:fig-confusion-matrix} 
\end{figure}

Example confusion matrices for four of the models on the topic classification task are shown in Fig. \ref{fig:fig-confusion-matrix}. The models shown are Linear SVM (Fig. \ref{fig:fig-conf-svm}), CNN model with regular word embeddings (Fig. \ref{fig:fig-conf-cnn}), CNN model with pre-trained word embeddings (Fig. \ref{fig:fig-conf-pretrain-cnn}), and BERT (Fig. \ref{fig:fig-conf-bert}). Any (hyper)parameters are the same as the ones for the results in the Table~\ref{tab:table-topics}. For reference, the topics and their distribution in the data are shown in Table \ref{tab:table-topics-data}. 

From Fig. \ref{fig:fig-conf-svm}, it seems that SVM did very well (91\%) on the larger topic (``Web Development", about 51\% of the records). BERT on the other hand seems to have done the best on the smaller topic (``Non-Programming") and relatively well (mid 70's) on the rest of the topics (see Fig. \ref{fig:fig-conf-bert}). CNN models did worse overall (see Figures \ref{fig:fig-conf-cnn} and \ref{fig:fig-conf-pretrain-cnn}), except for the CNN with pre-trained word embeddings matched the performance of the SVM for the large topic (``Web Development") (see Figures \ref{fig:fig-conf-svm} and \ref{fig:fig-conf-pretrain-cnn}).

\begin{table}
\caption{Comparison of Deep Learning models given different \textit{maxlen} values for Topic Classification - Runtime is Seconds per Epoch}
\centering
\begin{tabular}{lcccc}
\hline
\textit{\textbf{Model}}   & \textbf{\textit{maxlen}} &\textit{ \textbf{Runtime}} & ~~\textit{\textbf{Accuracy} }           & \textit{\textbf{F1-macro}}   \\ 
\hline
BERT    & 50     & 76       & 0.772 ~\small{±0.03} & 0.774 ~\small{±0.05} \\ 
BERT    & 100    & 114       & 0.799 ~\small{±0.03}  & 0.825 ~\small{±0.05}          \\ 
\hline
RoBERTa & 50     & 70       & 0.749 ~\small{±0.07} & 0.747 ~\small{±0.09} \\ 
RoBERTa & 100    & 125       & 0.794 ~\small{±0.04}      & 0.812 ~\small{±0.05}      \\
\hline
XLNet & 50     &   83   &  0.724 ~\small{±0.08} & 0.722 ~\small{±0.09}\\
XLNet & 100    &   164   &   0.757 ~\small{±0.05}   & 0.776 ~\small{±0.05}   \\
\hline

Word-Based CNN     & 50     & 8     & 0.645 ~\small{±0.04} & 0.623 ~\small{±0.07} \\ 
Word-Based CNN     & 100    & 10    & 0.727 ~\small{±0.04} & 0.733 ~\small{±0.06}             \\ 
\hline
\end{tabular}
	\label{tab:table-topics-maxlen}
\end{table}

As we did for the sentiment analysis task (see section \ref{sec:results-senti}), we also experimented with higher \textit{maxlen} values for the topic-based classification task. Results are shown in Table \ref{tab:table-topics-maxlen}. For \textit{maxlen} set to 100, BERT surpassed the Linear SVM result: BERT's F1-macro was 82.5\% versus the 80.6\% of the SVM (shown in Table \ref{tab:table-topics}). 
However, BERT achieved this at around 2 minutes per epoch, which then made the 5-fold stratified CV and 3 epoch execution at about half an hour. 

\begin{figure}[t] 	
\begin{subfigure}{0.45\textwidth}
         \centering
\includegraphics[scale=0.35]{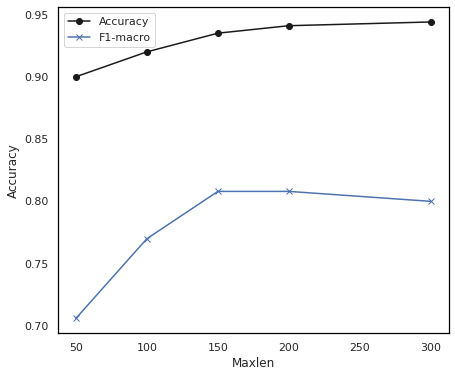}\\ 
\caption{Sentiment Analysis}
\end{subfigure}
\hfill
\begin{subfigure}{0.45\textwidth}
\centering
\includegraphics[scale=0.35]{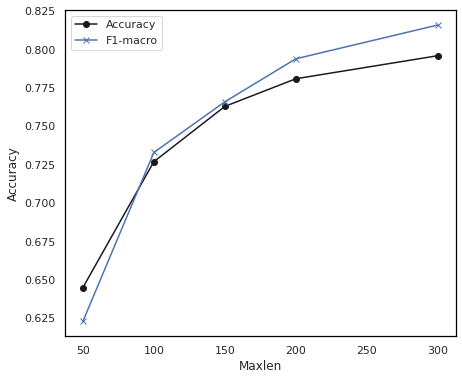}
\caption{Topic Classification}
\end{subfigure}
\caption{Effect of \textit{maxlen} on Accuracy and F1-macro for the CNN Model (regular word embeddings) for both tasks} 
\label{fig:maxlen} 
\end{figure}

As the CNN models are the two best of the regular DL models, we also experimented with increasing \textit{maxlen} for the CNN. Fig. \ref{fig:maxlen}b shows the effect of different \textit{maxlen} values on the classification performance of the CNN model with regular word embeddings for the topic-based classification task. As can be seen in Fig. \ref{fig:maxlen}, as \textit{maxlen} increased (which means that more of the text from the comment is used as input to the model), the accuracy and F1-macro overall tended to increase, but the increase was much more pronounced for the topic classification task (Fig. \ref{fig:maxlen}b) than the sentiment analysis task (Fig. \ref{fig:maxlen}a). For \textit{maxlen} set at 300, the CNN model performed as well as the Linear SVM to classify the course topics (see Table \ref{tab:table-topics}). It is noteworthy that we did not observe a similar effect on the accuracy of the linear SVM by limiting number of tokens or features as the input parameter, and that the SVM has a smaller vocabulary than the CNN model. We conducted similar experiments for the LSTM and we did not see an increase in accuracy. As a note, in these results, the CV experiments led to a range of ±0.05 to ±0.07 deviation from the numbers shown in the plots.

\begin{figure}[t] 	
\begin{subfigure}{0.45\textwidth}
         \centering
\includegraphics[scale=0.35]{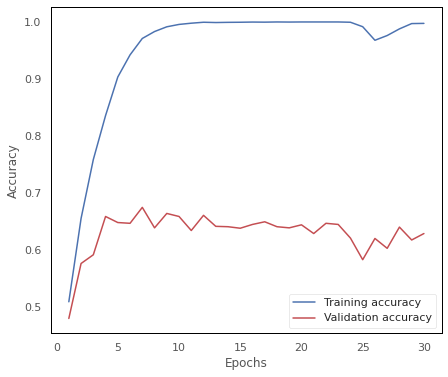}
\caption{Accuracy}
     \end{subfigure}
\hfill
     \begin{subfigure}{0.45\textwidth}
        \centering
\includegraphics[scale=0.35]{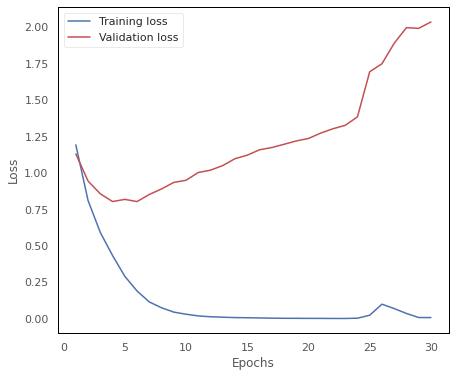}
\caption{Loss}
\end{subfigure}
\caption{Accuracy and Loss per epoch for the CNN Model (pre-trained word embeddings) on Topic-Based Classification} 	
\label{fig:fig-epochs-cnn} 
\end{figure}

Finally, we also investigated the effect of the number of epochs for the CNN model. Fig. \ref{fig:fig-epochs-cnn} shows an example of a typical run using the pre-trained word-based CNN on the topic classification task. As can be seen from both the accuracy and loss figures, the best values for the number of epochs were 4-6, after which the network started overfitting on the training data.

Overall, we see that the topic classification task is more challenging for the models than the sentiment analysis. This was expected as the binary sentiment analysis task has been shown to be relatively more straightforward for DL models in recent years. 
Our empirical results also indicate that we should use a larger part of each review (in the form of the \textit{maxlen} parameter) to feed as input into the DL models to improve performance. 
After manually exploring several reviews chosen at random, we observed that sometimes the topic-related wording did not appear until later in the comment. This verified that using a larger part of the course review would indeed result in better model performance. Finally, some of the reviews might lack topic-related terms or language, or they might have wording that is shared by more than one topic, and therefore they are more difficult to classify even by a human. For example, given the following snippets of a `Web Development' course review: 
\textit{``This class is a good jump start into a technical career! The class has a limited amount of time and it's really hard to go over as much content as there is, but it's all necessary to get down the fundamentals in that timeframe. [...] is willing to help people with coding problems after it is over [...]''}: this review could easily be classified into `Programming' instead of `Web Development' (note that none of the parts we omitted from this example included any terms or language specific to web development). 

A limitation of this work is the way we selected and assigned the topics: for example, the `Non-Programming' topic included reviews for courses in Digital marketing or UX Design, which made the topic not as well-focused as the rest of the topics. At the same time, the `Programming' topic included reviews for courses in iOS, Android, or Full Stack Development, which might also have very different terminology. Finally, as we just discussed, the terms or expressions used in a review for a `Programming' course could very well apply to a review for a `Web development' course. As our dataset are available to others, future research could look into different topics that are more detailed or assigned differently. 

\section{Conclusions}
\label{sec:conclusions}
In this study, we describe how we collected and pre-processed more than
ten thousand  course review comments publicly
available online. We present extensive experimentation
with several ML techniques to extract
sentiment from the text in the reviews as well as to detect the topic of the course for which the review was written. 
The techniques with which we experiment included a traditional bag-of-words
representation of the text as well as word embeddings and
character embeddings. Our employed classification models
range from traditional machine learning, such as Naïve
Bayes and SVMs, to current DL techniques, based
on CNNs and LSTMs. Finally, we fill a gap in the current research by exploring state-of-the-art transformer-based models (BERT \citep{devlin-etal-2019-bert}, RoBERTa \citep{liu2019roberta}, and XLNet \citep{yang2019xlnet}) which have not been used extensively yet in this course review analysis field. 

Our extensive experimentation with these algorithms shows how the different models behave for the two tasks. For the sentiment analysis task, the state-of-the-art transformer-based NLP models perform the best. For the topic classification task, the traditional models, such as SVMs, perform the best, though the DL models become top-performing when we increased the fraction of the course review that is fed as input to the model (using a hyperparameter called \textit{maxlen}). At the same time, we provide a complete picture by showing how the state-of-the-art models require much longer execution times to achieve their results. 

Sentiment analysis and topic classification can be used by educators and administrators as part of their assessment process in order to continuously improve instruction delivery and address issues. Our empirical results, exploration, and discussion can serve to guide others in the analysis of their own course feedback data. Our data and models could be used by others in their own course feedback analysis. 
Future research goals are to further explore our data towards aspect-based sentiment analysis. We also plan to explore other features in the data, such as helpfulness of the
reviews, and explore the use of additional pre-trained models such as EduBERT \citep{clavie2019edubert}. Finally, we would like to explore the applicability of our pre-trained DL models on other student feedback data.



\subsection*{Declarations}

\textbf{Availability of data and materials:} The datasets generated during and/or analysed during the current study are available from the corresponding author on reasonable request.

\noindent \textbf{Conflict of interest:} The authors declare that there is no conflict of interest regarding the publication of this paper and that the work presented in this article is not supported by any funding agency.



\bibliography{references}


\end{document}